\documentclass[conference]{article}
\usepackage[margin=3cm]{geometry}
\usepackage{cite}
\usepackage{gensymb}
\usepackage{amsmath,amssymb,amsfonts}
\usepackage{algorithmic}
\usepackage{graphicx}
\usepackage{textcomp}
\usepackage{tabularx, longtable, booktabs, multirow, tabularray}
\usepackage{xcolor}

\title{U-NetMN and SegNetMN: Modified U-Net and SegNet models for bimodal SAR image segmentation}

\author{%
Marwane Kzadri
\textit{LISAC} \\
\textit{Sidi Mohamed Ben Abdellah Univ}\\
Fez, Morocco \\
marwane.kzadri@usmba.ac.ma
\and
Franco Alberto Cardillo \\
\textit{Institute for Computational Linguistics} \\
\textit{National Research Council}\\
Pisa, Italy \\
francoalberto.cardillo@cnr.it
\and
Nanée Chahinian
\textit{HSM} \\
\textit{IRD, Univ Montpellier, CNRS}\\
Montpellier, France \\
nanee.chahinian@ird.fr
\and
Carole Delenne
\textit{IUSTI} \\
\textit{Aix Marseille Univ., CNRS}\\
Marseille, France \\
carole.delenne@univ-amu.fr
\and
Renaud Hostache
\textit{Espace Dev} \\
\textit{Univ. Montpellier, IRD}\\
Montpellier, France \\
renaud.hostache@ird.fr
\and
Jamal Riffi
\textit{LISAC} \\
\textit{Sidi Mohamed Ben Abdellah Univ}\\
Fez, Morocco \\
riffijamal@gmail.com
}

\date{}

\DeclareMathOperator*{\argmax}{argmax}

\begin{document}

\maketitle

\begin{abstract}
Segmenting Synthetic Aperture Radar (SAR) images is crucial for many remote sensing applications, particularly water body detection. However, deep learning-based segmentation models often face challenges related to convergence speed and stability, mainly due to the complex statistical distribution of this type of data. In this study, we evaluate the impact of mode normalization on two widely used semantic segmentation models, U-Net and SegNet. Specifically, we integrate mode normalization, to reduce convergence time while maintaining the performance of the baseline models. Experimental results demonstrate that mode normalization significantly accelerates convergence. 
Furthermore, cross-validation results indicate that normalized models exhibit increased stability in different zones. These findings highlight the effectiveness of normalization in improving computational efficiency and generalization in SAR image segmentation.
\end{abstract}

\section{Introduction}

Flood monitoring and water body segmentation are crucial tasks in remote sensing, particularly for disaster management and environmental monitoring. Synthetic Aperture Radar (SAR) imagery is widely used because of its ability to penetrate clouds and operate under all weather conditions, making it highly suitable for various remote sensing applications.

Traditional techniques of water segmentation based on thresholding or classical machine learning algorithms such as Support Vector Machines (SVM) and Random Forests (RF) struggle with generalization across different SAR datasets. Recently, deep learning approaches such as U-Net~\cite{ronneberger2015} and SegNet~\cite{Badrinarayanan2017} have demonstrated promising results, but they often suffer from sensitivity to dataset imbalance and lack robustness when applied to different regions. Moreover, these models require a significant amount of training time due to their complex architectures and the reliance on batch normalization (BN)~\cite{Ioffe2015}, which assumes a unimodal distribution of activations. This assumption can lead to a suboptimal convergence speed, particularly in datasets with multimodal distributions.

To address this limitation, we propose the integration of Mode Normalization (MN)~\cite{Deecke2018}, which dynamically adapts to multimodal data distributions. Unlike BN, MN allows the model to converge faster by reducing the instability caused by heterogeneous feature distributions, thus optimizing computational efficiency during training. By improving convergence speed, MN has the potential to significantly reduce training time while maintaining or even improving segmentation accuracy.

\section{Related Works}\label{sec:relatedworks}
In recent years, the application of deep learning architectures, particularly U-Net, has significantly advanced the segmentation of water bodies in Synthetic Aperture Radar (SAR) imagery~\cite{Ye2024}. These frameworks leverage convolutional neural networks (CNNs) structured to address challenges posed by pixel-level classifications in remote sensing tasks, particularly in distinguishing water features from complex backgrounds.
U-Net has been widely adopted for water body detection because of its encoder-decoder architecture, which enables high-resolution spatial information retention during the down-sampling and up-sampling processes. Zhang et al.~\cite{Zhang2023} developed an automated method using U-Net for accurate shoreline extraction from high-resolution SAR data, achieving precision and recall rates of 0.8 and 0.9, respectively. This underscores U-Net's effectiveness in complex environments~\cite{Chang2024}.

The segmentation of water bodies using SegNet has also been reinforced by studies that emphasize the model's robustness in various contexts. According to Lv et al.~\cite{Lv2022}, CNNs, including SegNet, have been essential to improve the accuracy of flood mapping in SAR images, highlighting their applicability in monitoring dynamic hydrological events. These findings indicate that deep learning models are increasingly preferred to traditional thresholding and statistical methods, particularly in complex scenarios where conventional techniques struggle with shadow effects and surface variability.

Normalization in neural networks is a crucial preprocessing step that improves the efficiency of training and the predictive performance of these models.
Normalization standardizes input data to reduce issues like internal covariate shift, where data distributions change between layers during training. Wang et al.~\cite{Wang2024} indicate that normalization techniques, such as scaling by mean and variance, significantly improve neural network performance, notably improving accuracy in tasks such as incident detection.

Batch normalization has emerged as a widely adopted technique that stabilizes the learning process by normalizing the inputs of each layer~\cite{Ioffe2015,Kohler2018}. This method not only accelerates training, but also enhances the overall accuracy of deep learning architectures~\cite{Kohler2018,Jin2015}. However, it may not translate well across multimodal data~\cite{Darma2021}.The functional interdependencies among modalities further complicate BN~\cite{Wang2024}.

\begin{table}[t!]
\centering
\caption{Main attributes of the SAR imagery used in this study.}
\label{tab:attributs}
\begin{tabular}{|l|l|} 
\hline
\textbf{Attribute} & \textbf{Value}                               \\ 
\hline
Satellite          & Sentinel-1                                   \\ 
\hline
Image Dimensions   & 11,112$\times$6,706 pixels  \\ 
\hline
Pixel Size         & 20$\times$20 meters         \\ 
\hline
Data Type          & 32-bit floating-point (Float32)              \\ 
\hline
Backscatter Range  & -48.85 dB to 11.79 dB                        \\ 
\hline
Area Covered       & $\approx$ 30,000 km$^{2}$                       \\
\hline
\end{tabular}
\end{table}

\section{Materials}\label{sec:Materials}
\subsection{Dataset}\label{subsec:dataset}
Our study area is located near Santarém, Pará, Brazil, a region characterized by dynamic hydrological conditions, including water bodies and areas prone to flooding. Our raw dataset consists of two adjacent images captured by the Sentinel-1 (S1) satellite 25 seconds apart from each other. Due to the side-looking radar configuration of S1, its images provide a slanted view of the ground surface, resulting in two-dimensional raw data with regions along the edges containing no data. In our initial pre-processing stage, we merge the two images and label the areas containing the actual data that are to be processed.
The SAR images have a spatial resolution of 20~m. They are single-band images, whose values represent radar backscatter intensity measured in decibels (dB). In our dataset, the raw values range from  $-48.85$~dB to $11.79$~dB. Lower values correspond to smooth surfaces and, in particular, water, while higher values indicate rougher surfaces, such as buildings or vegetation. These variations in radar intensity can be displayed as grayscale images, where darker and lighter shades of gray correspond to lower and higher backscatter. The merged image devoid of no-data cells has a size of  11,112$\times$6,706 pixels, thus covering approximately 30,000 km$^{2}$. The main attributes of the images in our dataset are summarized in Table~\ref{tab:attributs}.

The mask provided with the dataset was generated using the Hierarchical Split-Based Approach (HSBA)~\cite{Chini2017} and verified by domain experts.

\begin{figure}[t!]
    \centering
    \includegraphics[width=0.75\linewidth]{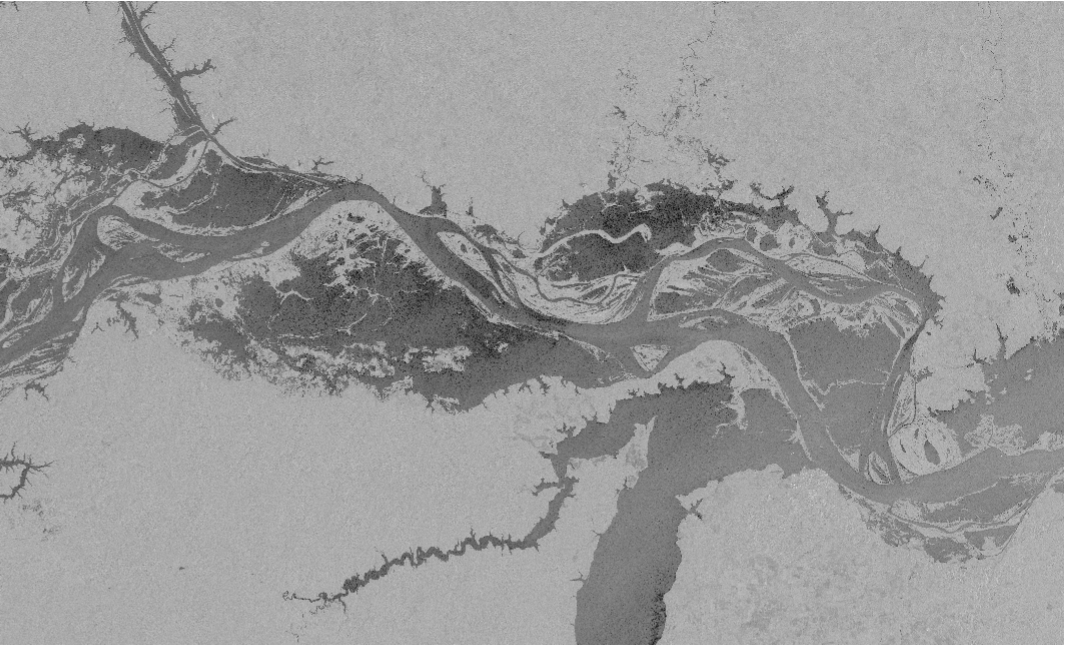}
    \caption{Study area near Santarém, Pará, Brazil.}
    \label{fig:largeimage}
\end{figure}

\begin{figure}[htbp]
    \centering
    \includegraphics[width=0.8\linewidth]{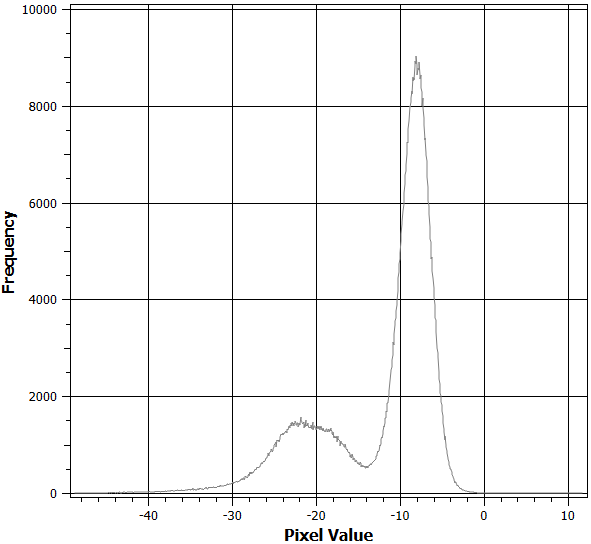}
    \caption{Histogram of the backscatter distribution in the SAR image.}
    \label{fig:histogram}
\end{figure}
\subsection{Preprocessing}

Following the original U-Net~\cite{ronneberger2015} and due to GPU memory constraints, we use a tiling strategy and partition the input image and mask into tiles of size 256$\times$ 256 pixels. The resulting dataset is composed of 1,118 tiles. In every experiment, we normalized the raw data $x$ using standardization $\hat{x} = \frac{x - \mu}{\sigma}$, 
where \( \mu \) and \( \sigma \) are, respectively, the mean and the standard deviation computed only on the subset used for training the model.

\subsection{Methods}\label{method}
This section introduces the models employed in our study, beginning with the U-Net model, followed by the SegNet model, and concluding with the mode normalization approach.

\subsubsection{U-Net}
The U-Net model is widely recognized for its effectiveness in segmentation tasks. Proposed by Ronneberger et al.~\cite{ronneberger2015} for biomedical image segmentation, U-Net follows an encoder-decoder architecture with two symmetrical paths. The encoder consists of two convolutional layers followed by 2$\times$2 max-pooling operations, where each step doubles the number of feature channels. This process reduces spatial dimensions while capturing high-level semantic information. The decoder reverses this contraction by applying transposed convolutions (upconvolution), which increase the spatial resolution and restore the input dimensions. Additionally, skip connections concatenate feature maps from the encoder to the decoder at corresponding levels, preserving spatial details. This process is followed by 2$\times$2 convolutional layers, which refine the features and gradually reconstruct the original resolution. The final layer is a 1$\times$1 convolution with a number of filters equal to the required segmentation classes.
\subsubsection{SegNet}
SegNet is an encoder-decoder architecture. The encoder, inspired by the first 13 convolutional layers of the VGG16 network~\cite{Simonyan2014}, extracts features from the input image. Each encoder layer performs a series of operations: convolution, batch normalization, ReLU activation, and max-pooling, progressively reducing spatial dimensions while enriching feature representation~\cite{Badrinarayanan2017}.
The SegNet decoder, which mirrors the encoder, also comprises 13 corresponding layers. Its mission is to reconstruct a segmented image from the extracted features~\cite{Badrinarayanan2017}. To counteract the loss of boundary details due to max-pooling, SegNet memorizes max-pooling indices. These indices are used during upsampling in the decoder, allowing precise reconstruction of boundaries, crucial for high-quality segmentation.
Each decoder layer performs upsampling using memorized indices, followed by convolution and batch normalization. This process is repeated until feature maps of the same size as the input of the corresponding encoder are obtained. The final layer applies a sigmoid activation, providing pixel-wise classification into the target class or the background.
\subsubsection{Our proposal}
Our dataset has a bimodal distribution, as clearly shown in the histogram plotted in Fig.~\ref{fig:histogram}. The presence of two modes in the data led us to experiment with a new normalization strategy, called Mode Normalization (MN), introduced by Deecke et al.~\cite{Deecke2018}
to improve model training with datasets characterized by a multimodal distribution. This approach detects multiple data modes in the input distribution and extends normalization of the input samples to multiple means and variances leading, as we will show, to reduced training times without loss in performance. 

A conventional Batch Normalization (BN) layer performs two distinct operations. First, it normalizes the activations $x_i$ of each mini-batch of size $m$ to have zero mean and unit variance:
\begin{align}
    \mu_B = & \frac{1}{m} \sum_{i=1}^{m} x_i & \quad \sigma_B^2 &=  \frac{1}{m} \sum_{i=1}^{m} (x_i - \mu_B)^2 \\
    \hat{x}_i = & \frac{x_i - \mu_B}{\sqrt{\sigma_B^2 + \epsilon}} \label{eq:bn1}\\
\intertext{where $\epsilon$ is a small value to avoid a possible division by zero in case of null variance.
Then, it applies a learnable affine transformation to the normalized activations $\hat{x}_i$ :}
 \quad y_i = & \gamma \hat{x}_i + \beta \label{eq:bn2}
\end{align}
where $\beta$ and $\gamma$ are two parameters optimized during training and $y_i$ are the 
activations forwarded to the successive layer in the neural network.

Mode Normalization extends Batch Normalization by considering a mixture of multiple distributions, capturing different data modes. Instead of computing a single mean and variance, MN detects \( K \) modes (K=2 in our case) in the data, which is assumed to be drawn from a mixture of \( K \) Gaussian distributions, parameterized by:
   \begin{equation}
       P(x) = \sum_{k=1}^{K} \pi_k \mathcal{N}(x \mid \mu_k, \sigma_k^2)
   \end{equation}
where \( \pi_k \) are mixture weights such that \( \sum_{k} \pi_k = 1 \), learned during training~\cite{Deecke2018}, \( \mu_k \) and \( \sigma_k^2 \) are the mean and variance of the \( k \)-th mode.

   Each sample \( x_i \) is assigned to the mode $k^*$ with the highest posterior probability $P(k^* \mid x_i)$:
\begin{align}
       k^* & = \argmax_{k \in \{1 \ldots K\}} P(k \mid x_i) \\
\intertext{The Bayes' theorem enables to compute the posterior probability:}
       P(k \mid x_i) & = \frac{\pi_k \mathcal{N}(x_i \mid \mu_k, \sigma_k^2)}{\sum_{j} \pi_j \mathcal{N}(x_i \mid \mu_j, \sigma_j^2)}
\end{align}
Once the sample is assigned to the mode $k^*$, normalization is performed similarly to BN as in Eqs.~\eqref{eq:bn1} and~\eqref{eq:bn2}:

    \begin{equation}
       \hat{x}_i = \frac{x_i - \mu_{k^*}}{\sqrt{\sigma_{k^*}^2 + \epsilon}} ; \qquad
       y_i = \gamma_{k^*} \hat{x}_i + \beta_{k^*}
   \end{equation}

with \( \gamma_{k^*} \) and \( \beta_{k^*} \) being mode-specific learnable parameters~\cite{Deecke2018}.

\section{Experiment}\label{experiment}
We follow the approach described in~\cite{Fernandez2014} and structure our experimentation in two stages. In the first stage (Sec.~\ref{subsec:hyper-search}), we optimize the hyperparameters of the models. To do this, we randomly partition the dataset into a single training/validation/test split using stratified sampling and evaluate the performance of the models for the various configurations of the hyperparameters. In the second stage (Sec.~\ref{subsec:cv}), we evaluate the generalization ability of the models using k-fold cross-validation. In this phase, we use the optimal hyperparameter values computed in the first stage. This approach has been criticized for potential test data contamination between the two stages~\cite{Wainberg2016}. Although we acknowledge this limitation, we accepted the trade-off due to the high computational cost and long training times of the models, given our available resources, and due to the limited dataset. We leave more robust and theoretically sound experiments for future work. The performance of the models has been evaluated using the metrics described in the next section.

\subsection{Evaluation Metrics}\label{subsec: metrics}
The performance of the segmentation models is evaluated using the following metrics:

\begin{equation}\label{eq:metrics}
\begin{aligned}
    &\text{Accuracy}   & = \quad & \mathrm{\frac{TP + TN}{TP + TN + FP + FN}} \\[0.5em]
    &\text{Precision}  & = \quad  & \mathrm{\frac{TP}{TP + FP}  }               \\[0.5em]
    & \text{Recall}     & = \quad  & \mathrm{\frac{TP}{TP + FN} } \\[0.5em]
    & \text{F1-Score}  & = \quad  & \frac{2 \times \text{Precision} \times \text{Recall}}{\text{Precision} + \text{Recall}} \\[0.5em]
    & \text{IoU}      & = \quad & \mathrm{\frac{TP}{TP + FP + FN}} \\[0.5em]
    & \text{Dsc}      & = \quad  & \mathrm{\frac{2\times TP}{2 \times TP + FP + FN} } \\
\end{aligned}
\end{equation}
where, in binary classification, TP (true positive) and FP (false positive) are the numbers of correctly and incorrectly classified positive instances. TN (true negative) and FN (false negative) are the same for the negative class. In our task, the positive class corresponds to \emph{``water''} and the negative class to \emph{``non-water''}. IoU and Dsc indicate the Intersection over Union and the Dice Similarity Coefficient, respectively.

\begin{table}[t!]
\centering
\caption{Hyperparameter Grid }\label{tab:Hyperparam}
\begin{tabular}{|l|l|}
\hline
\textbf{Hyperparameter}           & \textbf{Set of values}                         \\ \hline

Optimizer & Adam, SGD                \\ \hline
Learning Rate            & $10^{-4}, 10^{-3}, 10^{-2}$            \\ \hline

Dropout Rate             & $0.0,0.1,0.2,0.3,0.5$          \\ \hline
Loss Function & Dice, Focal, Combined loss            \\ \hline
\end{tabular}
\end{table}

\subsection{Hyperparameter Search }\label{subsec:hyper-search}
Hyperparameters are configuration variables that are set prior to training. They are not learned from the data, but they have indeed a strong influence on the training process. Their tuning is an important step in optimizing the final performance of the model~\cite{Bhandari2024,Ilemobayo2024}. We performed an exhaustive search on the hyperparameter grid shown in Table~\ref{tab:Hyperparam} for the two baseline models U-Net and SegNet. Specifically, we evaluated the influence of different optimizers, learning rates, loss functions and dropout rates. The Combined Loss integrates Dice Loss and Focal Loss to balance class distributions and enhance segmentation performance~\cite{Azad2023}. 
\begin{align}
    \mathcal{L}_{\text{Dice}} &= 1 - \frac{2 \sum y_{\text{true}} y_{\text{pred}} + \epsilon}{\sum y_{\text{true}} + \sum y_{\text{pred}} + \epsilon} \\
    \mathcal{L}_{\text{Focal}}(p_t) &= -\alpha_t (1 - p_t)^{\gamma} \log(p_t)
     \\
    \mathcal{L}_{\text{combined}} &= 0.5 \times \mathcal{L}_{\text{Dice}} + 0.5 \times \mathcal{L}_{\text{Focal}}
\end{align}
We trained and evaluated the two baseline models using a single random partition of the dataset into training (70\%), validation (10\%) and test (20\%) sets built via stratified sampling, and performed several training runs for each model. The two sets of hyperparameter values leading the two models to the best Dsc value (Eq.~\eqref{eq:metrics}) on the validation set are used in the second stage for training both the baseline and the MN-based derived model.

\subsection{Cross-Validation}\label{subsection: quarter experiment}
We performed a 4-fold cross-validation (CV) by dividing the image in Fig.~\ref{fig:largeimage} and the label mask into four non-overlapping zones, corresponding to image quarters. At each iteration of the CV, we used three zones as training set and the remaining one as test set. After four iterations, each zone was used once as test set and three times as part of the training set.

\section{Results}\label{sec:result}
In this section, we present the results obtained from the two experiments: the hyperparameter search and the zone-based cross-validation. In all the experiments, we used a batch size of 32 and applied early stopping with a patience of five epochs on the model loss, meaning that the training is stopped if the validation loss has not decreased during the last five epochs.

\subsection{Hyperparameter Search}
Table~\ref{tab:HyperparamValues} summarizes the optimal hyperparameters identified for the U-Net and SegNet models based on the Dsc value over the test set. We preferred Dsc over IoU as it is more sensitive to class imbalance and gives a better measure of performance when the segmentation involves small areas.

\begin{table}[t!]
\centering
\caption{Optimal hyperparameter values} 
\begin{tblr}{
  row{1} = {c},
  cell{1}{1} = {r=2}{},
  cell{1}{2} = {c=4}{},
  vlines,
  hline{1,3-5} = {-}{},
  hline{2} = {2-5}{},
}\label{tab:HyperparamValues}
\textbf{Model} & \textbf{Parameters} &                        &                  &                        \\
               & \textbf{optimizer}  & \textbf{leraning rate} & \textbf{dropout} & \textbf{loss function} \\
U-Net          & Adam                &  $10^{-4}$                & 0.1              & Dice loss              \\
SegNet         & Adam                & $10^{-3}$                 & 0               & Dice loss              
\end{tblr}
\end{table}

We fixed the best hyperparameters and trained both models for a maximum of 60 epochs applying early breaking based on patience on the validation loss, as previously described. Table~\ref{tab:results} shows the results obtained by the four models over the test set.

\begin{table}[h!]
\centering
\caption{Test metrics in the hyperparameter grid search }
\label{tab:results}
\begin{tblr}{
  row{1} = {c},
  cell{1}{1} = {r=2}{},
  cell{1}{2} = {c=6}{},
  vlines,
  hline{1,3-7} = {-}{},
  hline{2} = {2-7}{},
}
\textbf{Model} & \textbf{Metric}   &                    &                 &                   &              &              \\
               & \textbf{Accuracy} & \textbf{Precision} & \textbf{Recall} & \textbf{F1-Score} & \textbf{IoU} & \textbf{Dsc} \\
U-Net          & 0.991            & 0.984              & 0.986           & 0.984             & 0.970        & 0.984        \\
SegNet         & 0.961            &  0.925              & 0.937          &0.929             &  0.872      & 0.931       \\
U-NetMN        & 0.989             & 0.982              &0.979           & 0.980             & 0.963        & 0.980        \\
SegnetMN       &  0.949             & 0.908              &  0.914           & 0.909            & 0.838        & 0.911        
\end{tblr}
\end{table}

\subsection{Cross validation}\label{subsec:cv} 
Table~\ref{tab:crossvalidation} presents the metrics in Eq.~\eqref{eq:metrics} for the 4-fold CV experiments.

\begin{figure}[htbp]
    \centering
    \includegraphics[width=0.9\linewidth]{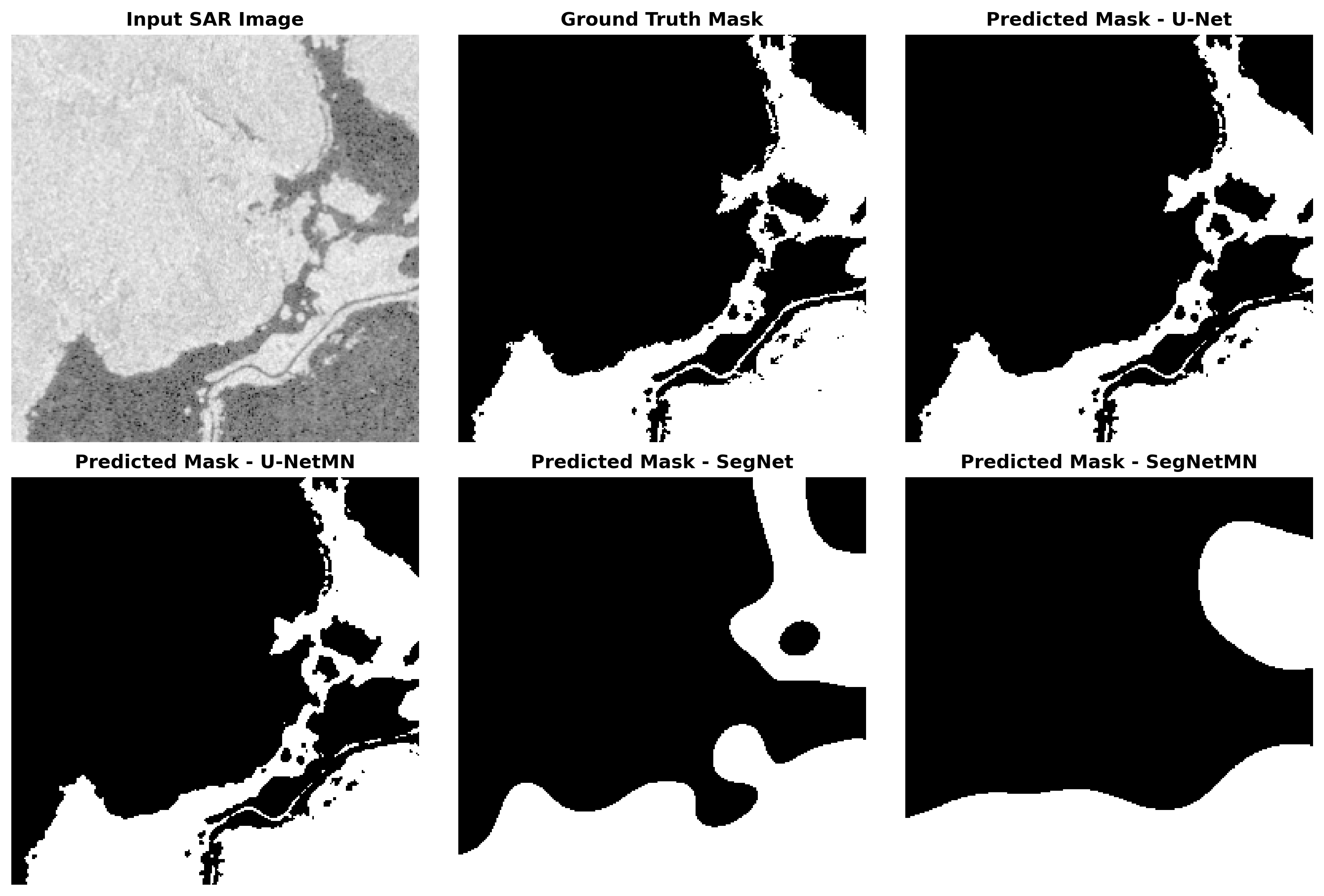}
    \caption{Example of segmentation. Clockwise from the top left: input tile, ground-truth,  U-Net, U-NetMN, SegNet, and SegNetMN.}
    \label{fig:segmentation_results}
\end{figure}

\begin{figure}[htbp]
    \centering
    \includegraphics[width=0.9\linewidth]{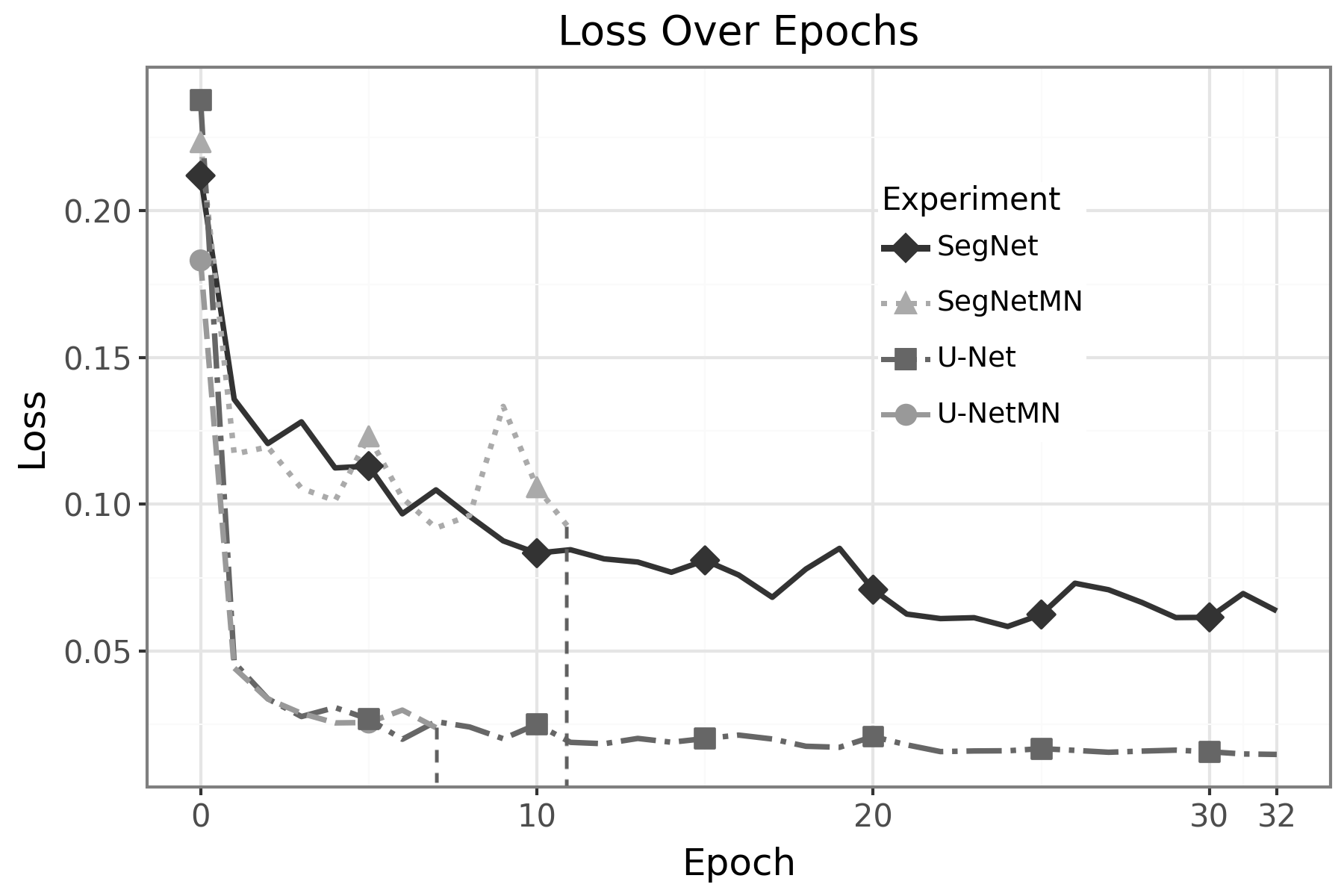}
    \caption{Loss curves for U-Net, SegNet, U-NetMN, and SegNetMN during training.}
    \label{fig:loss curve}
\end{figure}

\begin{table}[htbp]
\centering
\caption{4-fold Cross Validation Results}

\label{tab:crossvalidation}
\begin{tblr}{
  cell{1}{1} = {r=2}{},
  cell{1}{2} = {r=2}{},
  cell{1}{3} = {c=4}{c},
  cell{3}{1} = {r=6}{},
  cell{3}{3} = {c},
  cell{3}{4} = {c},
  cell{3}{5} = {c},
  cell{3}{6} = {c},
  cell{4}{3} = {c},
  cell{4}{4} = {c},
  cell{4}{5} = {c},
  cell{4}{6} = {c},
  cell{5}{3} = {c},
  cell{5}{4} = {c},
  cell{5}{5} = {c},
  cell{5}{6} = {c},
  cell{6}{3} = {c},
  cell{6}{4} = {c},
  cell{6}{5} = {c},
  cell{6}{6} = {c},
  cell{7}{3} = {c},
  cell{7}{4} = {c},
  cell{7}{5} = {c},
  cell{7}{6} = {c},
  cell{8}{3} = {c},
  cell{8}{4} = {c},
  cell{8}{5} = {c},
  cell{8}{6} = {c},
  cell{9}{1} = {r=6}{},
  cell{9}{3} = {c},
  cell{9}{4} = {c},
  cell{9}{5} = {c},
  cell{9}{6} = {c},
  cell{10}{3} = {c},
  cell{10}{4} = {c},
  cell{10}{5} = {c},
  cell{10}{6} = {c},
  cell{11}{3} = {c},
  cell{11}{4} = {c},
  cell{11}{5} = {c},
  cell{11}{6} = {c},
  cell{12}{3} = {c},
  cell{12}{4} = {c},
  cell{12}{5} = {c},
  cell{12}{6} = {c},
  cell{13}{3} = {c},
  cell{13}{4} = {c},
  cell{13}{5} = {c},
  cell{13}{6} = {c},
  cell{14}{3} = {c},
  cell{14}{4} = {c},
  cell{14}{5} = {c},
  cell{14}{6} = {c},
  cell{15}{1} = {r=6}{},
  cell{15}{3} = {c},
  cell{15}{4} = {c},
  cell{15}{5} = {c},
  cell{15}{6} = {c},
  cell{16}{3} = {c},
  cell{16}{4} = {c},
  cell{16}{5} = {c},
  cell{16}{6} = {c},
  cell{17}{3} = {c},
  cell{17}{4} = {c},
  cell{17}{5} = {c},
  cell{17}{6} = {c},
  cell{18}{3} = {c},
  cell{18}{4} = {c},
  cell{18}{5} = {c},
  cell{18}{6} = {c},
  cell{19}{3} = {c},
  cell{19}{4} = {c},
  cell{19}{5} = {c},
  cell{19}{6} = {c},
  cell{20}{3} = {c},
  cell{20}{4} = {c},
  cell{20}{5} = {c},
  cell{20}{6} = {c},
  vlines,
  hline{1,3,9,15,21} = {-}{},
  hline{2} = {3-6}{},
  hline{4-8,10-14,16-20,22-26} = {2-6}{},
}
\textbf{Model} & \textbf{Metric} & \textbf{Test Set}  &                 &                 &                 \\
               &                 & \textbf{Zone 1} & \textbf{Zone} 2 & \textbf{Zone} 3 & \textbf{Zone 4} \\
U-Net          & Accuracy        & 0.989               & 0.995               & 0.993              & 0.985               \\
               & Precision       & 0.986               & 0.985               & 0.976               & 0.977               \\
               & Recall          &  0.983              & 0.981               &  0.976               & 0.990               \\
               & F1-Score        & 0.984               & 0.974              & 0.979               & 0.983               \\
               & IoU             &  0.969               & 0.967              &   0.961              & 0.968               \\
               & Dsc             & 0.984               & 0.982               & 0.979              & 0.983               \\
SegNet         & Accuracy        & 0.915               & 0.961              &  0.857               &0.954               \\
               & Precision       &  0.843              & 0.802              & 0.554              & 0.948               \\
               & Recall          & 0.927               & 0.952               & 0.958              & 0.948               \\
               & F1-Score           &0.879              &  0.856               & 0.689              & 0.947               \\
               & IoU             & 0.791               &0.772               & 0.544               & 0.901              \\
               & Dsc             & 0.882              & 0.905              &  0.699               & 0.948              \\
U-NetMN        & Accuracy        &  0.984               & 0.995               & 0.991               & 0.985               \\
               & Precision       & 0.973               & 0.989               & 0.980              &  0.977               \\
               & Recall          & 0.981               & 0.973               & 0.970               & 0.989               \\
               & F1-Score           & 0.976               & 0.965               & 0.972               & 0.982               \\
               & IoU             & 0.955               & 0.962               & 0.952               & 0.967               \\
               & Dsc             & 0.976               &  0.980               & 0.973   
               & 0.982     \\

SegNetMN         & Accuracy        & 0.955               & 0.971               &  0.946               & 0.963               \\
               & Precision       &  0.922              & 0.896             & 0.899               & 0.970               \\
               & Recall          &0.918               & 0.906               & 0.851             & 0.908              \\
               & F1-Score           &  0.917              & 0.890               & 0.867               &  0.937              \\
           & IoU             & 0.852               & 0.815              &  0.777               &  0.884              \\
               & Dsc             & 0.920               &  0.896              &  0.873               & 0.938               \\
               \hline
\end{tblr}
\end{table}
Fig.~\ref{fig:segmentation_results} presents the input image, the ground truth and the segmentation masks predicted by different models. Both U-Net and U-NetMN produce segmentation masks that closely resemble the benchmark mask supplied with the images. Table ~\ref{tab/epoch and times conv} presents the number of epochs required to train each, along with the corresponding training time.
\begin{table}
\centering
\caption{Training epochs and convergence time for different models}
\label{tab/epoch and times conv}
\begin{tabular}{|l|c|c|c|} 
\hline
\textbf{Model} & \multicolumn{1}{l|}{\textbf{Training Epochs}} & \multicolumn{1}{l|}{\textbf{Training Time (s)}} & \textbf{Speed-up} \\ 
\hline
U-Net          & 33                                               & 320                                   & 1                \\ 
\hline
SegNet         & 32                                               & 227                                   & 1                \\ 
\hline
U-NetMN        & 8                                                & 167                                   & $\approx$ 1.92                \\ 
\hline
SegNetMN       & 12                                               & 123                                   & $\approx$ 1.85                \\
\hline
\end{tabular}
\end{table}
Fig.~\ref{fig:loss curve} presents the loss curves of the four models. The plotted curves clearly illustrate that the training of models with Mode Normalization stopped much earlier than those without. Specifically, the training of the U-Net model was stopped at epoch 33 after 320 seconds, whereas for U-NetMN it was early-stopped at epoch eight after 167 seconds. Similarly, the training of SegNetMN was stopped earlier than SegNet, with only a slight difference in performance between the two models. These results highlight the effectiveness of normalization in accelerating convergence and stabilizing training.

As shown in Table~\ref{tab:crossvalidation}  the final performance of the U-Net and U-NetMN models is nearly identical. However, there is a significant difference in computational efficiency. The standard U-Net model converges after 33 epochs out of 60, requiring approximately 320 seconds. In contrast, the U-Net model with mode normalization (U-NetMN) converges in just 8 epochs in 167 seconds, achieving the same accuracy in less than a quarter of the training time.

The SegNet model also achieves competitive results, with negligible differences compared to SegNetMN. However, the incorporation of normalization significantly improves the convergence speed. SegNetMN reaches convergence in approximately 123 seconds (12 epochs), whereas the original SegNet model requires around 227 seconds (32 epochs).

Regarding the cross-validation experiment, SegNetMN demonstrates relatively stable performance across all zones, with variations ranging for precision, IoU and Dsc from 2\% to 11\%. In contrast, the original SegNet model exhibits substantial performance fluctuations, losing nearly 30\% in Zone 3 comparing the highest recorded accuracy in Zone 4. 

The results obtained in this study confirm our initial hypothesis: with bimodal data, the Mode Normalization reduces convergence time while preserving the performance of the original models. Furthermore, with Mode Normalization the models seem to reach a more stable performance in the four different zones, despite the variations in the distribution of water pixels within the images.  The significant reduction in convergence time is particularly remarkable for both models used in this study (U-Net and SegNet), lowering from 25 epochs compared to the original model. Normalization also helped maintain model stability across application areas for the SegNet model. This enables the model to be more robust to variations in the characteristics of SAR images.
      
\section{Conclusion}
In this study, we investigated two semantic segmentation models and explored the integration of mode normalization to reduce the training time on bimodal SAR images, while maintaining the performance of the baseline models. The key findings demonstrate that U-Net and SegNet with mode normalization converge faster than the original models. Moreover, normalization improves the stability of the model in different cross-validation zones. 
Although the results are promising, we acknowledge some limitations of our work. First, the experimental settings introduce contamination between training and test data. Second, the hyperparameters' optimization involved just a handful of parameters and only the two base models. Third and last, as we used a single image, we cannot conclude anything about the generalization of the models. All of these major limitations will be tackled in our future work, which will be based on a larger dataset that we are currently collecting.

\section*{Acknowledgment}

This research has received support from the European Union's Horizon research and innovation program under the MSCA-SE (Marie Sk\l{}odowska-Curie Actions Staff Exchange) grant agreement 101086252; Call: HORIZON-MSCA-2021-SE-01; Project title: STARWARS (STormwAteR and WastewAteR networkS heterogeneous data AI-driven management).

Experiments presented in this paper were carried out using the Grid'5000 testbed, supported by a scientific interest group hosted by Inria and including CNRS, RENATER and several Universities and organizations (see https://www.grid5000.fr).

\end{document}